\documentclass[letterpaper,10pt,conference]{IEEEtran}
\usepackage{graphicx}
\usepackage{courier}
\usepackage{cite}
\usepackage{bm}
\usepackage{nccmath}
\usepackage{amssymb}
\usepackage{indentfirst}
\usepackage{algorithm}
\usepackage[hidelinks,pagebackref]{hyperref}
\usepackage{algpseudocode}
\usepackage{orcidlink}

\newcommand{\mat}[1]{\bm{\mathrm{#1}}}
\newcommand{\tr}{\mathrm{tr}}

\makeatletter
\newcommand{\mathleft}{\@fleqntrue\@mathmargin0pt}
\newcommand{\mathcenter}{\@fleqnfalse}
\makeatother

\algnewcommand\algorithmicnot{\textbf{not}}
\algdef{SE}[IF]{IfNot}{EndIf}[1]{\algorithmicif\ \algorithmicnot\ #1\ \algorithmicthen}{\algorithmicend\ \algorithmicif}%
\algnewcommand\algorithmicforeach{\textbf{for each}}
\algdef{S}[FOR]{ForEach}[1]{\algorithmicforeach\ #1\ \algorithmicdo}

\begin{document}

\title{\LARGE\textbf{MultiDLO: Simultaneous Shape Tracking of Multiple Deformable Linear Objects with Global-Local Topology Preservation}}

\author{Jingyi Xiang\textsuperscript{1}\orcidlink{0000-0003-0727-3098}, Holly Dinkel\textsuperscript{2}\orcidlink{0000-0002-7510-2066}%
\thanks{\textsuperscript{1}Department of Electrical and Computer Engineering and Coordinated Science Laboratory, University of Illinois at Urbana-Champaign, Urbana, IL, 61801. \texttt{jingyix4@illinois.edu.}}
\thanks{\textsuperscript{2}Department of Aerospace Engineering and Coordinated Science Laboratory, University of Illinois at Urbana-Champaign, Urbana, IL, 61801. \texttt{hdinkel2@illinois.edu.}}
}

\maketitle

\begin{abstract} 

MultiDLO is a real-time algorithm for estimating the shapes of multiple, intertwining deformable linear objects (DLOs) from RGB-D image sequences. Unlike prior methods that track only a single DLO, MultiDLO simultaneously handles several objects. It uses the geodesic distance in the Global-Local Topology Preservation algorithm to define both inter-object identity and intra-object topology, ensuring entangled DLOs remain distinct with accurate local geometry. The MultiDLO algorithm is demonstrated on two challenging scenarios involving three entangling ropes, and the implementation is open-source and available for the community.

\end{abstract}
\IEEEpeerreviewmaketitle

\section{Introduction}
\label{introduction}

Consider an automated robotic system which monitors in real-time the shape of a deformable linear object (DLO), for example a rope, a wire, or a string. This system could perceive the DLO in RGB-D imagery and estimate its configuration to perform a closed-loop manipulation task such as shape control or wire routing, or it could monitor the DLO for collision prevention  \cite{yan2020supervised, yan2020TMP, lagneau2020shapecontrol, yin2021domanipulation, yu2022shapecontrol, yu2023shapecontrol, jin2022routing}. These tasks are common in applications like robotic surgery, industrial automation, power line avoidance and human habitat maintenance. Previous work used physics simulation, including Finite Element Method (FEM) analysis, multi-physics, and dynamics, to model DLO motion \cite{schulman2013deformable,ruan2018rigidity,zhang2021llldynamics}, and motion planning frameworks predict minimal-energy wire configurations using the wire tips as boundary conditions \cite{moll2006energy,bretl2015kinematic,bretl2015kirchoff}. This work builds on existing tracking methods through tracking multiple deforming and entangling DLOs simultaneously. This work make the following contributions:

\begin{enumerate}
    \item MultiDLO is a method of tracking the shape of multiple DLOs in real-time. This is achieved by using the geodesic distance in the kernel describing how pairs of nodes influence each other’s motion and setting the distance between independent objects to infinity.
    \item MultiDLO demonstrates tracking of multiple DLOs without instance segmentation in each frame. This is achieved by performing instance segmentation (expensive) on the first frame for initialization and performing semantic segmentation (cheap) on subsequent frames.
    \item The source code and demonstration data are openly released at: \\ \href{https://github.com/RMDLO/multidlo}{https://github.com/RMDLO/multidlo}.
\end{enumerate}

\begin{figure}
\centering
\includegraphics[width=\columnwidth]{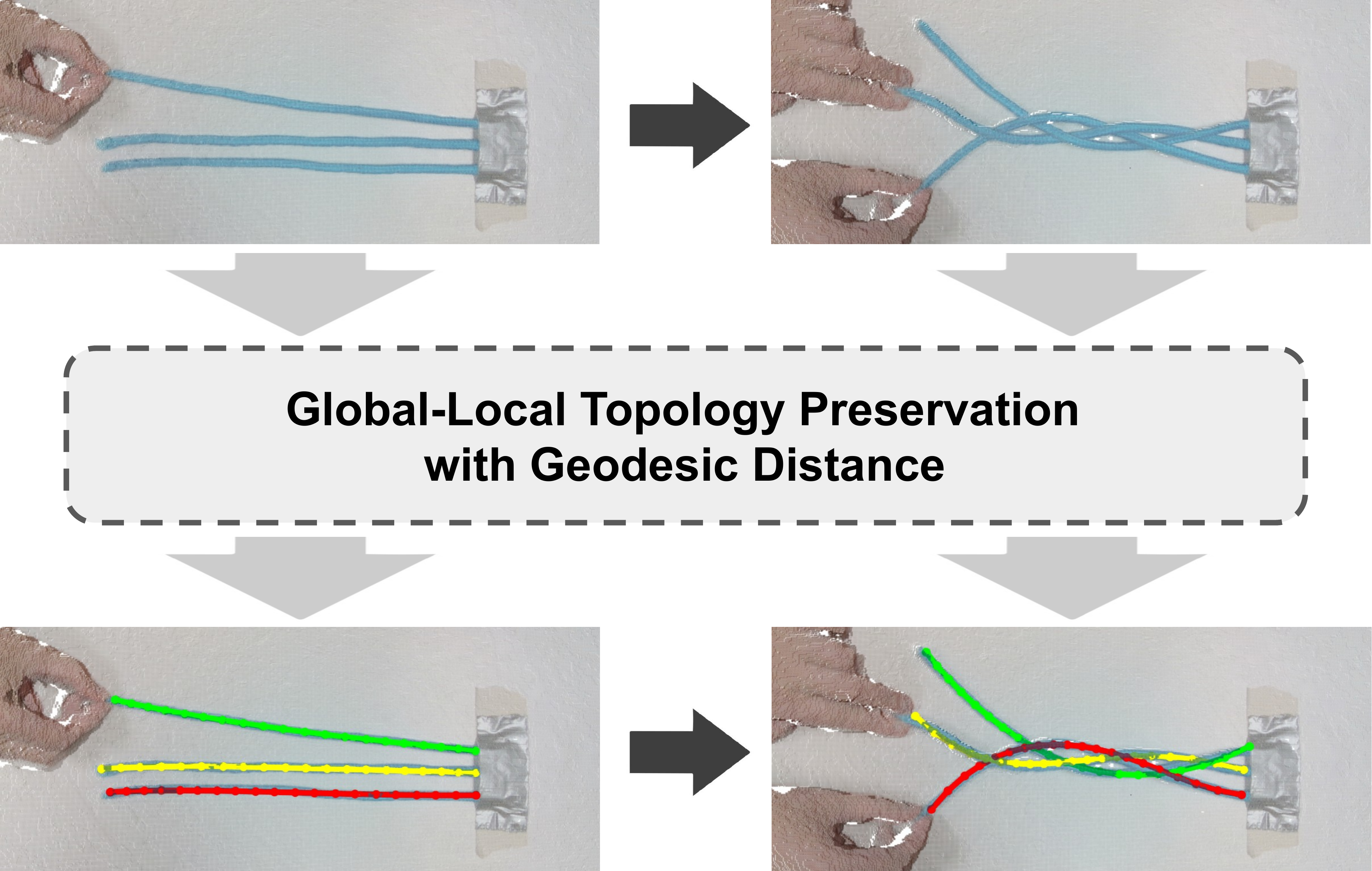}
\caption{Given a sequence of frames, MultiDLO uses Global-Local Topology Preservation with the geodesic distance metric to track the shape of multiple deformable linear objects simultaneously.}
\label{fig:intro}
\end{figure}

\section{Related Work}

The Global-Local Topology Preservation (GLTP) algorithm performs modified non-rigid point set registration to map one set of points onto another using the Expectation-Maximization (EM) framework proposed by the Coherent Point Drift (CPD) algorithm \cite{myronenko2006point,myronenko2010point,ge2014gltp}. The objective function for EM in GLTP uses CPD with locally linear embedding to preserve local topology. Non-rigid point set registration from CPD and GLTP set the foundation for several algorithms which perform DLO tracking under occlusion, including CPD+Physics, Structure Preserved Registration, CDCPD, CDCPD2, and TrackDLO \cite{tang2017state,tang2018framework,tang2022track, chi2019cdcpd, wang2021cdcpd2, xiang2023trackdlo}. These methods only demonstrate tracking one DLO at a time.

\section{Methods}

For $N$ \emph{points} in $\mathbb{R}^3$ received at time $t$ from a depth sensor, $\mat{X}^t_{N \times 3} = (\mat{x}_1^t, \hdots, \mat{x}_N^t)^T$, the DLO shape can be represented by a collection of $M$ ordered \emph{nodes}, $\mat{Y}^t_{M \times 3} = (\mat{y}_1^t, \hdots, \mat{y}_M^t)^T$. The shape of each of $K$ DLOs in a scene is represented by a different set of nodes, $\mat{Y}_{k\in [1,K]}^t$, each consisting of $M_k$ nodes. Stacking all the $\mat{Y}_k^t$ matrices vertically produces $\mat{Y}^t$ with shape $M \times 3$, where $M = \sum_{k=1}^K M_k$. All $K$ DLOs are treated as one combined object in the tracking process. From the tracking output $\mat{Y}^t$, the individual DLO nodes are obtained by accessing the corresponding rows in $\mat{Y}^t$.


Single DLO shape tracking algorithms generally perform object instance segmentation (conventionally, through color thresholding) in the RGB image first, and then segment the point cloud based on the RGB image segmentation \cite{chi2019cdcpd, wang2021cdcpd2}. Our approach only requires instance segmentation for the first frame to initialize $\mat{Y}^t$. By treating all DLOs as one deformable object, only semantic segmentation is required in subsequent frames. This approach bypasses the limitation of DLO instance segmentation in 2D RGB images, which can be slow and perform poorly \cite{zanella2021autogenerated,caporali2022fastdlo,caporali2022ariadneplus,dinkel2022RMDLO}.

\subsection{Gaussian Mixture Model (GMM) Node Registration}

\begin{figure}
\centering
\includegraphics[width=\columnwidth]{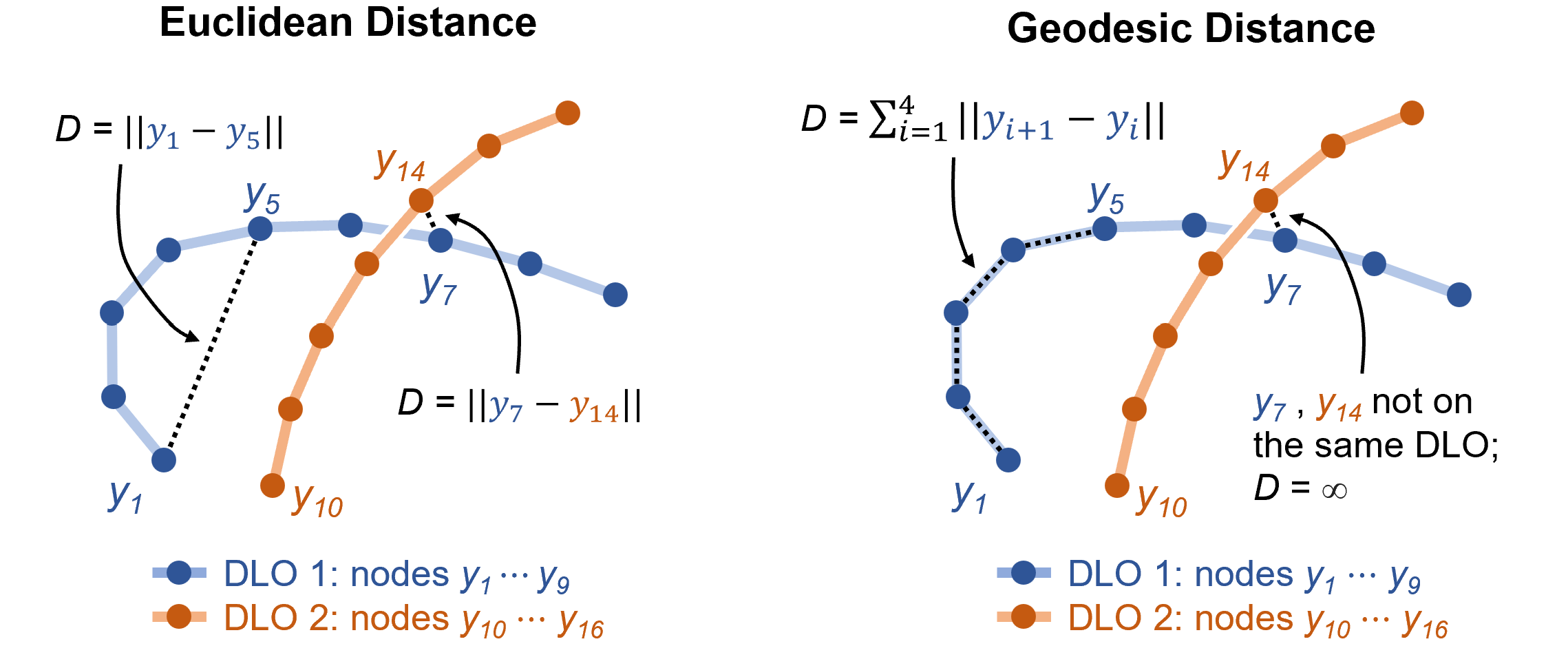}
\caption{Using Euclidean distance, the distance between nodes is the Euclidean norm of their difference. Using geodesic distance, the distance between two nodes on the same DLO is set to the sum of the segment lengths between them. If the two nodes are not on the same DLO, the distance between them is set to infinity.}
\label{fig:geodesic-proximity}
\end{figure}

Tracking begins with Gaussian Mixture Model (GMM) clustering as performed in \cite{tang2017state}. The GMM clustering step computes $\mat{Y}^t$ as the centroids of Gaussian distributions from which $\mat{X}^t$ points are randomly sampled with isotropic variance $\sigma^2$. Assuming equal membership probability $p(m) = \frac{1}{M}$ and $\mu$ percent of the points are outliers, the GMM cost function takes the form

\begin{equation} \label{eq:E-GMM-ignored-consts}
    \begin{split}
        & E_{\mathrm{GMM}}(\mat{y}^t_m, \sigma^2) \\
        & \ = \sum_{n=1}^{N} \sum_{m=1}^{M} p(m {\mid} \mat{x}^t_n) \frac{\|\mat{x}^t_n - \mat{y}^t_m \|^2}{2\sigma^2} + \frac{3 N_p}{2} \log (\sigma^2),
    \end{split}
\end{equation}
where
\begin{gather}
    N_p = \sum_{n=1}^{N} \sum_{m=1}^{M} p(m {\mid} \mat{x}^t_n) \\
    p(m|\mat{x}^t_n)
        = \frac{\exp \left( \frac{-D_{\mat{y}^t_m, \mat{x}^t_n}^2}{2\sigma^2} \right)}{\sum_{m=1}^{M} \exp \left( \frac{-D_{\mat{y}^t_m, \mat{x}^t_n}^2}{2\sigma^2} \right) + \frac{(2\pi \sigma^2)^{3/2}\mu M}{(1-\mu) N}}
\end{gather}
and $D$ is a distance metric (discussed in Section \ref{sec:geodesic-proximity}).

\subsection{The Motion Coherence Theory (MCT)}

Given node positions $\mat{Y}^{t-1}$ and $\mat{Y}^{t}$ from consecutive time steps, the MCT defines a spatial velocity field $v(\mat{Y}^{t-1}) = \mat{Y}^{t} - \mat{Y}^{t-1}$ \cite{yuille1989mathematical}. Nodes close to each other should move coherently through the smoothest possible spatial velocity field. For a spatial domain variable $\mat{z}$ and a frequency domain variable $\mat{s}$, the smoothness of the velocity field $v(\mat{z})$ can be measured by passing it through a high pass filter $1 / \Tilde{G}(\|\mat{s}\|)$ in the frequency domain as

\begin{equation} \label{eq:E-MCT}
    E_{\mathrm{MCT}}(v(\mat{z})) = \int_{\mathbb{R}^3} |\Tilde{v}(\mat{s})|^2 / \Tilde{G}(\|\mat{s}\|) d\mat{s},
\end{equation}

\noindent where $\Tilde{v}(\mat{s})$ and $\Tilde{G}(\|\mat{s}\|)$ are the Fourier Transforms of $v(\mat{z})$ and $G(\|\mat{z}\|)$. By choosing $G(\|\mat{z}\|) = \exp(-\frac{\|z\|^2}{2\beta^2})$, this cost term is equivalent to the cost term of the MCT. The parameter $\beta$ controls the width of high pass filter $1 / \Tilde{G}(\|\mat{s}\|)$. Therefore, a larger $\beta$ leads to a smoother velocity field.

\subsection{Locally Linear Embedding (LLE)}

Locally Linear Embedding (LLE) represents a node $\mat{y}^{t-1}_m$ with its closest $2Q$ neighbors and a set of weights $\mat{L}$. At the next time step $t$, LLE preserves local topology by reconstructing $\mat{y}_m^t$ with its new neighbors and $\mat{L}$:

\begin{equation}
    \begin{split}
        E_{LLE}(\mat{y}^{t}_m) = \sum_{m=1}^M \| \mat{y}^{t}_m - \sum_{i=m-Q}^{m+Q} \mat{L}(m, i) \mat{y}^{t}_i \|^2.
    \end{split}
\end{equation}
The weights $\mat{L}$ are computed by minimizing the cost~\cite{ghojogh2020locally},

\begin{equation}
    \begin{split}
        E_{LLE}(\mat{W}) = \sum_{m=1}^M \| \mat{y}^{t}_m - \sum_{i=m-Q}^{m+Q} \mat{L}(m, i) \mat{y}^{t}_i \|^2
    \end{split},
\end{equation}

\noindent with $\mat{W}$ encoded in $\mat{y}^t_m$ and $\mat{y}^t_i$ as 

\begin{equation}
    \begin{matrix}
    \mat{y}^t_m = \mat{y}^{t-1}_m + \mat{G}(m, \cdot)\mat{W} \\
    \mat{y}^t_i = \mat{y}^{t-1}_i + \mat{G}(i, \cdot)\mat{W}
    \end{matrix}.
\end{equation}

\subsection{Expectation-Maximization Update for GLTP}

The total cost of GLTP is updated iteratively using the EM algorithm. With $v(\mat{z})$ encoded in $\mat{y}_m^{t}$ as $\mat{y}_m^t = \mat{y}_m^{t-1} + v(\mat{y}_m^{t-1})$, the total cost is
\begin{equation} \label{eq:E-GMM-MCT-LLE}
    \begin{split}
        & E(v(\mat{z}), \sigma^2) = E_{\mathrm{GMM}} + E_{\mathrm{MCT}} + E_{\mathrm{LLE}}\\
        & = \sum_{n=1}^{N} \sum_{m=1}^{M} \frac{1}{2\sigma^2} p(m {\mid} \mat{x}^t_n) \|\mat{x}^t_n - \mat{y}_m^t \|^2 +  \frac{3 N_p}{2} \log (\sigma^2) \\
        & \ \ \ + \frac{\lambda}{2} \int_{\mathbb{R}^3} |\Tilde{v}(\mat{s})|^2 / \Tilde{G}(\|\mat{s}\|) d\mat{s} \\
        & \ \ \ + \sum_{m=1}^M \| \mat{y}^{t}_m - \sum_{i=m-Q}^{m+Q} \mat{L}(m, i) \mat{y}^{t}_i \|^2
    \end{split}.
\end{equation}\

\noindent The solution to the above cost function takes the form \cite{myronenko2010point}
\begin{equation}
    v(\mat{z}) = \sum_{m=1}^M \mat{w}_m G(\|\mat{z} - \mat{y}_m\|).
\end{equation}

\begin{figure*}
\centering
\includegraphics[width=\textwidth]{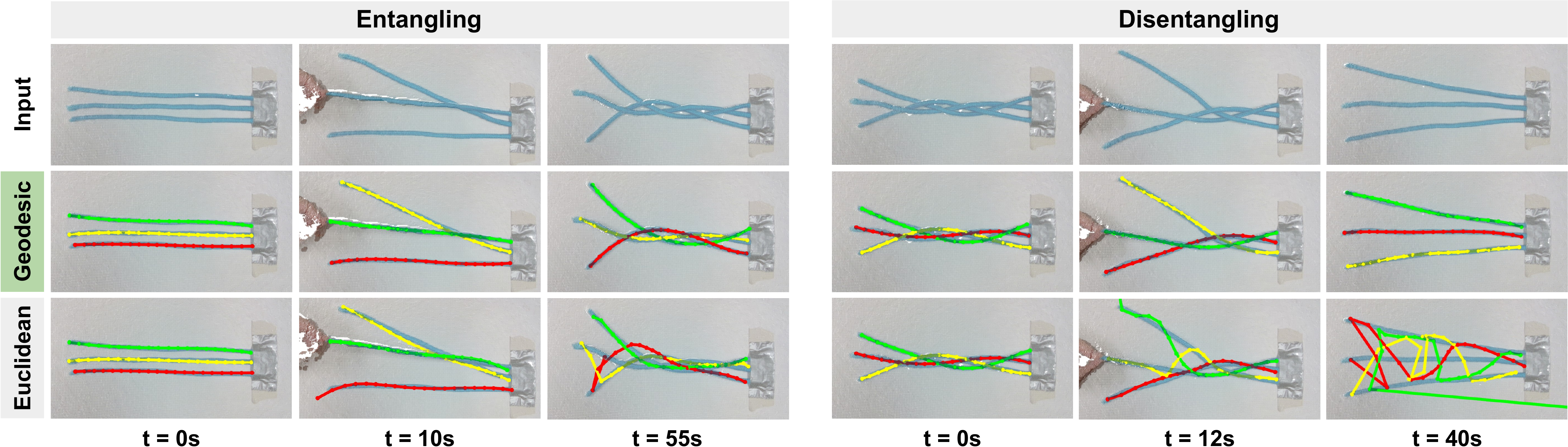}
\caption{Simultaneously tracking multiple DLOs with GLTP and geodesic proximity accurately estimates the shape of each DLO in demonstrations. By comparison, GLTP with Euclidean proximity fails to accurately track DLO shape in both the \textit{Entangling} and \textit{Disentangling} scenarios.}
\label{fig:comparison}
\end{figure*}

\noindent For readability, denote $\mat{X}^t$ as $\mat{X}$, $\mat{Y}^{t-1}$ as $\mat{Y}_0$, and use the following notations to solve for $\mat{w}_m$ and $\sigma^2$ analytically through EM:
\begin{itemize}
    \item $\mat{W}_{M \times 3} = (\mat{w}_1^t, \hdots, \mat{w}_M^t)^T$, kernel weights,
    \item $\mat{P}_{M \times N}$, posterior matrix with $\mat{P}(m, n) = p(m | \mat{x}_n^t)$,
    \item $\mat{G}_{M \times M}$, kernel matrix with $\mat{G}(i, m) = \exp(-\frac{D_{\mat{y}_i^t,\mat{y}_m^t}}{2\beta^2})$
    \item $\mat{H}_{M \times M} = (\mat{I}-\mat{L})^T (\mat{I}-\mat{L})$, LLE weights,
    \item $\mathrm{d}(\mat{a})$, the diagonal matrix constructed from vector $\mat{a}$,
    \item $\tr (\mat{m})$, the trace of matrix $\mat{m}$, and
    \item $\mat{1}$, a column vector of ones.
\end{itemize}

The solutions $\mat{W}$ and $\sigma^2$ are computed from taking the partial derivatives $\frac{\partial E}{\partial \mat{W}}$ and $\frac{\partial E}{\partial \sigma^2}$ and setting them to zero as
\mathleft
\begin{gather}
    \begin{split}
        \mat{W} = \ & (\mathrm{d}(\mat{P1}) \mat{G} + \lambda \sigma^2 \mat{I} + \alpha \sigma^2 \mat{HG})^{-1} \\
                  & \cdot (\mat{PX} - (\mathrm{d}(\mat{P1}) + \alpha \sigma^2 \mat{H}) \mat{Y}_0)
    \end{split} \\
    \begin{split}
        \sigma^2 = & \ \frac{1}{\mat{1}^T \mat{P1} D}(\tr(\mat{X}^T \mathrm{d}(\mat{P}^T \mat{1}) \mat{X}) - 2\tr((\mat{PX})^T \mat{Y}_0) \\
                   & + \tr(\mat{Y}_0^T \mathrm{d}(\mat{P1}) \mat{Y}_0) + \tr(\mat{W}^T \mat{G}^T \mathrm{d}(\mat{P1}) \mat{GW}) \\
                   & + 2\tr(\mat{W}^T \mat{G}^T \mathrm{d}(\mat{P1}) \mat{Y}_0) - \tr(\mat{W}^T \mat{G}^T \mat{PX}))
    \end{split}.
\end{gather}
\mathcenter

\noindent The new node positions are $\mat{Y}^{t} = \mat{Y}^{t-1} + \mat{GW}$.

\subsection{Geodesic Proximity}
\label{sec:geodesic-proximity}

The MCT requires the node velocity field to be smooth, where nodes spatially close to each other move in similar directions with similar speeds. In the node velocity given by
\begin{equation}
    \begin{split}
        & v(\mat{y}_m) \\
        & = \sum_{i = 1}^M \mat{G}(m, i) \mat{W}(i, \cdot)  = \sum_{i = 1}^M G(D_{\mat{y}_i, \mat{y}_m}) \mat{W}(i, \cdot),
    \end{split}
\end{equation} 
$D_{\mat{y}_i, \mat{y}_m}$ is the distance between nodes $\mat{y}_i$ and $\mat{y}_m$ given some distance metric $D$. Given $G$ is Gaussian, $G(D_{\mat{y}_i, \mat{y}_m})$ is small if $\mat{y}_i$ is far from $\mat{y}_m$. This produces a small $G(D_{\mat{y}_i, \mat{y}_m}) \mat{W}(i, \cdot)$, indicating the movement of $\mat{y}_i$ has insignificant influence on that of $\mat{y}_m$.

The Euclidean distance $d_{\mat{y}_i, \mat{y}_m}=\|\mat{y}_m - \mat{y}_i\|$ is a common choice for $D_{\mat{y}_i, \mat{y}_m}$, however it is not the best choice for representing the geometry of DLOs. If one DLO is resting on top of another DLO, the movement of the top one should have little influence on the movement of the bottom one. However, the small Euclidean distance between nodes near the intersection couples their motion together, causing tracking failure as shown in Fig. \ref{fig:comparison}.

To resolve this, MultiDLO uses geodesic distance for $D$ and defines the node-to-node geodesic distance $\rho_{\mat{y}_i,\mat{y}_m}$ to be
\begin{equation}
\label{eq:geodesic-node-to-node}
    \rho_{\mat{y}_i,\mat{y}_m} = 
    \begin{cases}
        \sum_{j=m}^{i-1} \|\mat{y}_{j+1} - \mat{y}_j\| & \text{if } m \leq i \\
        \sum_{j=i}^{m-1} \|\mat{y}_{j+1} - \mat{y}_j\| & \text{if } m > i
    \end{cases}.
\end{equation}
If $\mat{y}_i$ and $\mat{y}_m$ are from two different DLOs, the distance between them is set to infinity.

Similarly, two nodes close in Euclidean distance to an intersection but from different DLOs should have different proximities to points near the intersection. Where $\mat{y}_{c_1}$ and $\mat{y}_{c_2}$ are the two nodes closest to point $\mat{x}_n$, the node-to-point geodesic distance $\rho_{\mat{y}_m,\mat{x}_n}$ is

\begin{equation}
\label{eq:geodesic-distance}
    \rho_{\mat{y}_m,\mat{x}_n} = 
    \begin{cases}
        d_{\mat{y}_{c_1},\mat{x}_n} + \rho_{\mat{y}_{c_1},\mat{y}_m} & \text{if } \rho_{\mat{y}_{m},\mat{y}_{c_1}} \leq \rho_{\mat{y}_{m},\mat{y}_{c_2}} \\
        d_{\mat{y}_{c_2},\mat{x}_n} + \rho_{\mat{y}_{c_1},\mat{y}_m} & \text{if } \rho_{\mat{y}_{m},\mat{y}_{c_1}} > \rho_{\mat{y}_{m}, \mat{y}_{c_2}} \\
        d_{\mat{y}_{m},\mat{x}_n} & \text{if } m \in \{c_1, c_2\}
    \end{cases}.
\end{equation}

\section{Results}

Simultaneous MultiDLO shape tracking with intersection among independent objects is demonstrated. For these demonstrations, $Q = 3, \beta=0.8, \lambda=1$, $\alpha=3$, and the optimization function tolerance $\epsilon = 10^{-5}$. Performance is shown in two scenarios:

\begin{enumerate}
    \item \textit{Entangling}--Three ropes lie parallel on a table. The ropes are crossed over each other and all three become twisted. This scenario demonstrates tracking when DLOs are wound, knotted, or tied.
    \item \textit{Disentangling}--Three ropes lie in a twisted configuration on a table. The ropes are uncrossed until all three ropes reach a separated, parallel configuration. This scenario demonstrates tracking when DLOs are unwound, unknotted, or untied.
\end{enumerate}

The raw data comprise point cloud and RGB image data collected as the ropes are first wound and then unwound. The data are saved in one Robot Operating System (ROS) bag file. For initialization in both scenarios, instance segmentation was performed manually on the first frame due to the limitations of existing DLO instance segmentation algorithms \cite{caporali2022fastdlo,caporali2022ariadneplus,dinkel2022RMDLO}. The subsequent frames used color thresholding as the semantic segmentation input. The tracking results for these scenarios shown in Figure \ref{fig:comparison} highlight the accuracy of geodesic GLTP for multi-DLO shape tracking and the failure of Euclidean GLTP for this problem.

\section{Conclusions and Limitations}
\label{conclusions}

The MultiDLO algorithm is a real-time, accurate algorithm for tracking multiple DLOs as they are entangled. One limitation of the method is that a DLO could penetrate itself or other DLOs in the tracking result, which would not happen in real world collision situations. Potential solutions to this include incorporating physics simulators as described in Structure Preserved Registration or adapting the self-intersection constraint introduced in CDCPD2~\cite{tang2022track, wang2021cdcpd2}. This work could additionally be incorporated into a closed-loop multi-DLO shape controller for robotic manipulation.

\section*{Acknowledgements}

\noindent \small The authors thank the members of the Representing and Manipulating Deformable Linear Objects project (\href{https://github.com/RMDLO}{github.com/RMDLO}) and the teams developing and maintaining the open-source software used in this project \cite{ros, opencv_library, harris2020numpy, hunter2007matplotlib, zhou2018open3d, virtanen2020scipy, eigenweb, pcl}. This work was supported by the Illinois Space Grant Consortium and the NASA Space Technology Graduate Research Opportunity 80NSSC21K1292.

\bibliographystyle{IEEEtran}
\bibliography{IEEEabrv,references}

\begin{thebibliography}{10}
\providecommand{\url}[1]{#1}
\csname url@samestyle\endcsname
\providecommand{\newblock}{\relax}
\providecommand{\bibinfo}[2]{#2}
\providecommand{\BIBentrySTDinterwordspacing}{\spaceskip=0pt\relax}
\providecommand{\BIBentryALTinterwordstretchfactor}{4}
\providecommand{\BIBentryALTinterwordspacing}{\spaceskip=\fontdimen2\font plus
\BIBentryALTinterwordstretchfactor\fontdimen3\font minus
  \fontdimen4\font\relax}
\providecommand{\BIBforeignlanguage}[2]{{%
\expandafter\ifx\csname l@#1\endcsname\relax
\typeout{** WARNING: IEEEtran.bst: No hyphenation pattern has been}%
\typeout{** loaded for the language `#1'. Using the pattern for}%
\typeout{** the default language instead.}%
\else
\language=\csname l@#1\endcsname
\fi
#2}}
\providecommand{\BIBdecl}{\relax}
\BIBdecl

\bibitem{yan2020supervised}
M.~Yan, Y.~Zhu, N.~Jin, and J.~Bohg,
  ``\href{https://ieeexplore.ieee.org/document/8972568}{Self-Supervised
  Learning of State Estimation for Manipulating Deformable Linear Objects},''
  in \emph{{IEEE} Robot. Autom. Lett.}, vol.~5, no.~2, 4 2020, pp. 2372--2379.

\bibitem{yan2020TMP}
M.~Yan, G.~Li, Y.~Zhu, and J.~Bohg,
  ``\href{https://ieeexplore.ieee.org/document/9341330}{Learning Topological
  Motion Primitives for Knot Planning},'' in \emph{{IEEE/RSJ} Int. Conf.
  Intell. Robot. Sys. (IROS)}, 2020.

\bibitem{lagneau2020shapecontrol}
R.~Lagneau, A.~Krupa, and M.~Marchal,
  ``\href{https://ieeexplore.ieee.org/document/91330322}{Automatic Shape
  Control of Deformable Wires Based on Model-Free Visual Servoing},'' in
  \emph{{IEEE} Robot. Autom. Lett.}, vol.~5, 10 2020, pp. 5252--5259.

\bibitem{yin2021domanipulation}
H.~Yin, A.~Varava, and D.~Kragic,
  ``\href{https://www.science.org/doi/10.1126/scirobotics.abd8803}{Modeling,
  Learning, Perception, and Control Methods for Deformable Object
  Manipulation},'' in \emph{Sci. Rob.}, vol.~6, 5 2021, pp. 1--16.

\bibitem{yu2022shapecontrol}
M.~Yu, H.~Zhong, and X.~Li,
  ``\href{https://ieeexplore.ieee.org/document/9812244}{Shape Control of
  Deformable Linear Objects with Offline and Online Learning of Local Linear
  Deformation Models},'' in \emph{{IEEE} Int. Conf. Robot. Autom. (ICRA)}, 5
  2022, pp. 1337--1343.

\bibitem{yu2023shapecontrol}
M.~Yu, K.~Lv, C.~Wang, M.~Tomizuka, and X.~Li,
  ``{\href{https://ieeexplore.ieee.org/document/10160264}{A Coarse-to-Fine
  Framework for Dual-Arm Manipulation of Deformable Linear Objects with
  Whole-Body Obstacle Avoidance}},'' in \emph{{IEEE} Int. Conf. Robot. Autom.
  (ICRA)}, 5 2023.

\bibitem{jin2022routing}
S.~Jin, W.~Lian, C.~Wang, M.~Tomizuka, and S.~Schaal,
  ``\href{https://ieeexplore.ieee.org/document/9732654}{Robotic Cable Routing
  with Spatial Representation},'' \emph{{IEEE} Robot. Autom. Lett.}, vol.~7,
  no.~2, pp. 5687--5694, 2022.

\bibitem{schulman2013deformable}
J.~Schulman, A.~Lee, J.~Ho, and P.~Abbeel,
  ``\href{https://ieeexplore.ieee.org/abstract/document/6630714}{Tracking
  Deformable Objects with Point Clouds},'' in \emph{{IEEE} Int. Conf. Robot.
  Autom. (ICRA)}, 2013, pp. 1130--1137.

\bibitem{ruan2018rigidity}
M.~Ruan, D.~M$^{c}$Conachie, and D.~Berenson,
  ``\href{https://ieeexplore.ieee.org/document/8594520}{Accounting for
  Directional Rigidity and Constraints in Control for Manipulation of
  Deformable Objects without Physical Simulation},'' in \emph{{IEEE/RSJ} Int.
  Conf. Intell. Robot. Sys. (IROS)}, 2018, pp. 512--519.

\bibitem{zhang2021llldynamics}
W.~Zhang, K.~Schmeckpeper, P.~Chaudhari, and K.~Daniilidis,
  ``\href{https://ieeexplore.ieee.org/document/9560955}{Deformable Linear
  Object Prediction Using Locally Linear Latent Dynamics},'' in \emph{{IEEE}
  Int. Conf. Robot. Autom. (ICRA)}, 6 2021, pp. 13\,503--13\,509.

\bibitem{moll2006energy}
M.~Moll and L.~Kavraki,
  ``\href{https://ieeexplore.ieee.org/document/1668249/}{Path Planning for
  Deformable Linear Objects},'' \emph{{IEEE} Trans. Robot.}, vol.~22, pp.
  625--636, 2006.

\bibitem{bretl2015kinematic}
T.~Bretl and Z.~McCarthy,
  ``\href{https://ieeexplore.ieee.org/document/6327684}{Mechanics and
  Quasi-Static Manipulation of Planar Elastic Kinematic Chains},'' \emph{{IEEE}
  Trans. Robot.}, vol.~29, pp. 1--14, 2012.

\bibitem{bretl2015kirchoff}
------,
  ``\href{https://journals.sagepub.com/doi/abs/10.1177/0278364912473169}{Quasi-Static
  Manipulation of a Kirchoff Elastic Rod Based on a Geometric Analysis of
  Equilibrium Configurations},'' \emph{Int. J. Robot. Res.}, vol.~33, pp.
  48--68, 2013.

\bibitem{myronenko2006point}
A.~Myronenko, X.~Song, and M.~Carreira-Perpi\~{n}\'{a},
  ``\href{https://proceedings.neurips.cc/paper/2006/file/3b2d8f129ae2f408f2153cd9ce663043-Paper.pdf}{Non-Rigid
  Point Set Registration: Coherent Point Drift},'' \emph{Adv. Neur. Inf. Proc.
  (NeurIPS)}, pp. 1--8, 2006.

\bibitem{myronenko2010point}
A.~Myronenko and X.~Song,
  ``\href{https://ieeexplore.ieee.org/document/5432191}{Point Set Registration:
  Coherent Point Drift},'' in \emph{{IEEE} Trans. Pattern Anal. Mach. Intell.},
  vol.~32, no.~12, 12 2010, pp. 2262--2275.

\bibitem{ge2014gltp}
S.~Ge, G.~Fan, and M.~Ding,
  ``\href{https://ieeexplore.ieee.org/document/6909990}{Non-Rigid Point Set
  Registration with Global-Local Topology Preservation},'' \emph{{IEEE/CVF}
  Int. Conf. Comput. Vis. Pattern Recognit. Workshops (CVPRW)}, pp. 245--251,
  2014.

\bibitem{tang2017state}
T.~Tang, Y.~Fan, H.-C. Lin, and M.~Tomizuka,
  ``\href{https://ieeexplore.ieee.org/stamp/stamp.jsp?arnumber=8206058}{State
  Estimation for Deformable Objects by Point Registration and Dynamic
  Simulation},'' in \emph{{IEEE/RSJ} Int. Conf. Intell. Robot. Sys. (IROS)},
  2017, pp. 2427--2433.

\bibitem{tang2018framework}
T.~Tang, C.~Wang, and M.~Tomizuka,
  ``\href{https://ieeexplore.ieee.org/document/8403315}{A Framework for
  Manipulating Deformable Linear Objects by Coherent Point Drift},''
  \emph{{IEEE} Robot. Autom. Lett.}, vol.~3, no.~4, pp. 3426--3433, 2018.

\bibitem{tang2022track}
T.~Tang and M.~Tomizuka,
  ``\href{https://journals.sagepub.com/doi/full/10.1177/0278364919841431}{Track
  Deformable Objects from Point Clouds with Structure Preserved
  Registration},'' \emph{Int. J. Robot. Res.}, vol.~41, no.~6, pp. 599--614,
  2022.

\bibitem{chi2019cdcpd}
C.~Chi and D.~Berenson,
  ``\href{https://ieeexplore.ieee.org/document/8967827}{Occlusion-Robust
  Deformable Object Tracking Without Physics Simulation},'' in \emph{{IEEE/RSJ}
  Int. Conf. Intell. Robot. Sys. (IROS)}, 2019, pp. 6443--6450.

\bibitem{wang2021cdcpd2}
Y.~Wang, D.~M$^{c}$Conachie, and D.~Berenson,
  ``\href{https://ieeexplore.ieee.org/document/9561012}{Tracking
  Partially-Occluded Deformable Objects while Enforcing Geometric
  Constraints},'' in \emph{{IEEE} Int. Conf. Robot. Autom. (ICRA)}, 2021, pp.
  14\,199--14\,205.

\bibitem{xiang2023trackdlo}
J.~Xiang, H.~Dinkel, H.~Zhao, N.~Gao, B.~Coltin, T.~Smith, and T.~Bretl,
  ``\href{https://ieeexplore.ieee.org/document/10214157}{TrackDLO: Tracking
  Deformable Linear Objects Under Occlusion With Motion Coherence},''
  \emph{{IEEE} Robot. Autom. Lett.}, vol.~8, no.~10, pp. 6179--6186, 2023.

\bibitem{zanella2021autogenerated}
R.~Zanella, A.~Caporali, K.~Tadaka, D.~De~Gregorio, and G.~Palli,
  ``\href{https://ieeexplore.ieee.org/document/9349395}{Auto-Generated Wires
  Dataset for Semantic Segmentation with Domain Independence},'' in
  \emph{{IEEE} Int. Conf. Comput. Cont. Robot. (ICCCR)}, 1 2021, pp. 292--298.

\bibitem{caporali2022fastdlo}
A.~Caporali, K.~Galassi, R.~Zanella, and G.~Palli,
  ``\href{https://ieeexplore.ieee.org/document/9830852/}{FASTDLO: Fast
  Deformable Linear Objects Instance Segmentation},'' \emph{{IEEE} Robot.
  Autom. Lett.}, vol.~7, no.~4, pp. 9075--9082, 2022.

\bibitem{caporali2022ariadneplus}
A.~Caporali, R.~Zanella, D.~De~Gregorio, and G.~Palli,
  ``\href{https://ieeexplore.ieee.org/document/9721686}{Ariadne+: Deep
  Learning-Based Augmented Framework for the Instance Segmentation of Wires},''
  in \emph{{IEEE} Trans. Ind. Inf.}, 2 2022, pp. 1--11.

\bibitem{dinkel2022RMDLO}
H.~Dinkel, J.~Xiang, H.~Zhao, B.~Coltin, T.~Coltin, and T.~Bretl,
  ``\href{https://deformable-workshop.github.io/icra2022/spotlight/WDOICRA2022_08.pdf}{Wire
  Point Cloud Instance Segmentation from RGBD Imagery with Mask R-CNN},'' in
  \emph{{IEEE Int. Conf. Robot. Autom. (ICRA) Workshop on Representing and
  Manipulating Deformable Objects}}, 5 2022.

\bibitem{yuille1989mathematical}
A.~L. Yuille and N.~M. Grzywacz,
  ``\href{https://link.springer.com/article/10.1007/BF00126430}{A Mathematical
  Analysis of the Motion Coherence Theory},'' \emph{Int. J. Comput. Vis.},
  vol.~3, no.~2, pp. 155--175, 1989.

\bibitem{ghojogh2020locally}
B.~Ghojogh, A.~Ghodsi, F.~Karray, and M.~Crowley,
  ``\href{https://arxiv.org/abs/2011.10925}{Locally Linear Embedding and its
  Variants: Tutorial and Survey},'' \emph{arXiv preprint arXiv:2011.10925},
  2020.

\bibitem{ros}
{Stanford Artificial Intelligence Laboratory},
  ``{\href{https://www.ros.org}{Robotic Operating System: Noetic Ninjemys}},''
  https://www.ros.org, 2018.

\bibitem{opencv_library}
G.~Bradski, ``{The OpenCV Library},'' \emph{{Dr. Dobb's Journal of Software
  Tools}}, 2000.

\bibitem{harris2020numpy}
C.~R. Harris, J.~Millman, S.~van~der Walt, R.~Gommers, P.~Virtanen,
  D.~Cournapeau, E.~Wieser, J.~Taylor, S.~Berg, N.~J. Smith, R.~Kern, M.~Picus,
  S.~Hoyer, M.~H. van Kerkwijk, M.~Brett, A.~Haldane, J.~Fern\'{a}ndez~del
  R\'{i}o, M.~Wiebe, P.~Peterson, P.~G\'{e}rard-Marchant, K.~Sheppard,
  T.~Reddy, W.~Weckesser, H.~Abbasi, C.~Gohlke, and T.~E. Oliphant,
  ``\href{https://www.nature.com/articles/s41586-020-2649-2}{Array Programming
  with NumPy},'' \emph{Nature}, vol. 585, pp. 357--362, 2020.

\bibitem{hunter2007matplotlib}
J.~D. Hunter, ``\href{https://ieeexplore.ieee.org/document/4160265}{Matplotlib:
  A 2D Graphics Environment},'' \emph{Comput. Sci. Eng.}, vol.~9, no.~3, pp.
  90--95, 2007.

\bibitem{zhou2018open3d}
Q.-Y. Zhou, J.~Park, and V.~Koltun,
  ``\href{https://arxiv.org/abs/1801.09847}{Open3D: A Modern Library for 3D
  Data Processing},'' \emph{arXiv:1801.09847}, 2018.

\bibitem{virtanen2020scipy}
P.~Virtan, R.~Gommers, T.~E. Oliphant, M.~Haberland, T.~Reddy, D.~Cournapeau,
  E.~Burovski, P.~Peterson, W.~Weckesser, J.~Bright \emph{et~al.},
  ``\href{https://www.nature.com/articles/s41592-019-0686-2}{Scipy 1.0:
  Fundamental Algorithms for Scientific Computing in Python},'' \emph{Nat.
  Methods}, vol.~17, pp. 261–--272, 2020.

\bibitem{eigenweb}
G.~Guennebaud, B.~Jacob \emph{et~al.},
  ``\href{http://eigen.tuxfamily.org}{Eigen v3},'' http://eigen.tuxfamily.org,
  2010.

\bibitem{pcl}
R.~B. Rusu and S.~Cousins,
  ``\href{https://ieeexplore.ieee.org/document/5980567}{3D is Here: Point Cloud
  Library (PCL)},'' in \emph{{IEEE} Int. Conf. Robot. Autom. (ICRA)}, 2011, pp.
  1--4.

\end{thebibliography}
\end{document}